\documentclass[sigconf]{acmart}

\AtBeginDocument{%
  }

\usepackage{listings}
\usepackage{tabularx}
\usepackage{booktabs}
\usepackage{tabularx}
\usepackage{fontawesome5}
\usepackage{xcolor}
\usepackage{hyperref}  
\definecolor{productbrown}{RGB}{139, 69, 19}

\begin{document}


\title{Bridging Search and CRM: Productionizing AI Product Research Agents for Customer Re-Engagement}


\author{Mandar Kulkarni, Pooja A., Samir Shah}
\affiliation{%
  \institution{Flipkart Internet Pvt Ltd}
  \city{Bangalore}
  \country{India}}


\begin{abstract}

Modern e-commerce platforms often operate search, recommendation, personalization, and CRM systems independently, limiting opportunities for proactive customer re-engagement. This is particularly challenging for exploratory intents such as “best smartphones” or “latest 5G phones,” where users may leave the platform for external research before purchasing.
We present a scalable, production-deployed framework that bridges search and CRM workflows through AI-powered Product Research Agents. The system identifies users with exploratory purchase intent and low engagement, conducts grounded multi-agent product research using behavioral signals, external knowledge, and enterprise catalog data, and delivers personalized recommendations through WhatsApp.
We evaluate the framework in a 23-day production deployment involving approximately 15K WhatsApp notifications for mobile product discovery. The campaign achieved substantial CTR improvements over traditional WhatsApp recommendation campaigns, with evidence of secondary engagement through message forwarding and sharing. The deployment also generated downstream purchases and GMV impact, demonstrating the practical effectiveness of AI Product Research Agents for proactive customer re-engagement and end-to-end customer journey optimization.

\end{abstract}

\settopmatter{printacmref=false}
\renewcommand\footnotetextcopyrightpermission[1]{}  
\acmDOI{}
\acmISBN{}
\acmConference{}{}{}

\maketitle

\thispagestyle{fancy}
\fancyhf{}
\fancyhead[L]{\large Accepted at ACM SIGKDD 2026 5th Workshop on End-to-End Customer Journey Optimization}
\fancyfoot[C]{\thepage}

\section{Introduction}
\pagestyle{empty}

Modern e-commerce platforms increasingly seek to optimize the entire customer journey by integrating search, recommendation, personalization, and CRM systems into cohesive user experiences. However, these systems are often developed independently and optimized for isolated objectives such as click-through rate or short-term conversions, leading to fragmented customer interactions across discovery, research, engagement, and re-engagement stages. This limitation becomes particularly significant for exploratory and subjective purchase intents, where users require contextual understanding, comparative reasoning, and trustworthy recommendations before making purchasing decisions.

Subjective queries such as “best smartphones,” “latest 5G phones,” or “good mobiles for gaming” do not map cleanly to deterministic keyword-based retrieval systems. Addressing such queries requires synthesis of external knowledge, understanding of subjective preferences, awareness of evolving market trends, and generation of explainable recommendations. Conventional retrieval and ranking systems are primarily optimized for deterministic matching and relevance estimation, and therefore often return ranked product lists without contextual explanations or supporting evidence. Consequently, users frequently leave e-commerce platforms to conduct independent research through external channels such as YouTube reviews, web search engines, technology blogs, and community discussions before returning to complete purchases. This fragmented workflow introduces friction into the customer journey and reduces opportunities for sustained platform engagement.

To address this challenge, we present a scalable production-deployed enterprise framework that bridges search and CRM workflows using AI-powered Product Research Agents for proactive customer re-engagement. Rather than limiting optimization to in-session retrieval quality, the proposed framework identifies users exhibiting high-intent exploratory behavior but low engagement signals and proactively reconnects with them through personalized WhatsApp recommendations. The system combines large-scale behavioral analytics, grounded multi-agent reasoning, enterprise catalog integration, recommendation validation, and cross-channel CRM delivery into a unified end-to-end customer re-engagement pipeline.

At the core of the framework is a modular multi-agent architecture coordinated through a centralized orchestrator. A Query Analysis Agent first interprets subjective user intent by extracting structured signals such as product category, budget constraints, desired attributes, and latent preferences. A Discovery Agent subsequently performs external knowledge retrieval across web search, expert reviews, and community discussions to identify candidate products along with supporting evidence. The identified candidates are then grounded to internal commerce catalog entities using enterprise search APIs enriched with business constraints such as availability, delivery feasibility, personalized pricing, and discounts. To improve reliability and trustworthiness, the framework further incorporates a dedicated Review Agent responsible for validating technical specifications, release timelines, and factual consistency across sources, thereby reducing hallucinations and improving explainability.

A key design consideration of the proposed system is enterprise scalability. Instead of executing computationally expensive research workflows for all user traffic, the framework operates asynchronously over large-scale search query logs using a PySpark-based filtering pipeline. The pipeline identifies high-potential exploratory intents using signals such as zero-click behavior, affluence indicators, business relevance, and subjective query patterns. This selective processing framework enables efficient deployment of cost-intensive agentic reasoning pipelines at production scale while focusing interventions on impactful stages of the customer journey.

The generated recommendations are delivered through proactive WhatsApp (WA) notifications, enabling customer engagement beyond traditional in-app recommendation surfaces and allowing the platform to reconnect with potentially disengaged users outside the native commerce experience. In a live 23-day production deployment involving approximately 15K WhatsApp recommendation notifications for mobile product discovery, the proposed system achieved substantially higher click-through rates (\textasciitilde 285\%) compared to earlier WA mobile product campaign baselines. Multiple campaign days recorded click volumes exceeding the number of delivered notifications, suggesting strong secondary engagement driven by organic forwarding and message-sharing behavior. Furthermore, the deployment generated meaningful downstream purchasing activity and GMV impact, demonstrating the effectiveness of integrating scalable AI Product Research Agents with CRM-driven customer re-engagement workflows for end-to-end customer journey optimization in real-world enterprise commerce systems.

\begin{figure*}[!t]
  \centering
  \includegraphics[width=\textwidth]{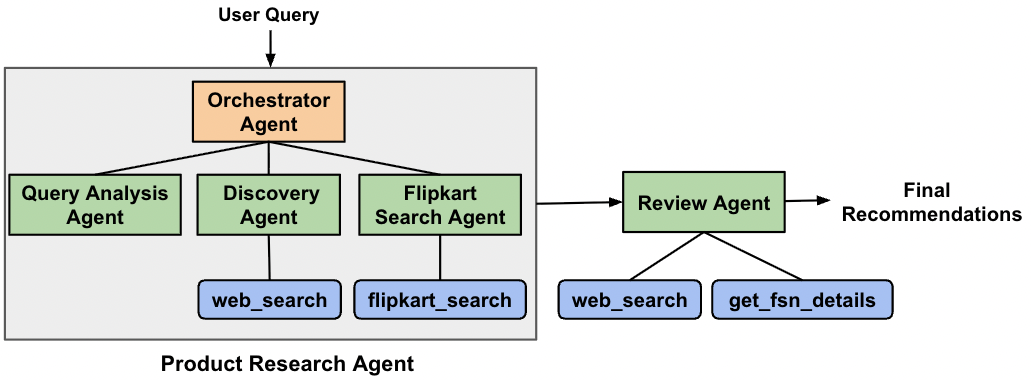}
  \caption{Proposed architecture of the Research Agent for Product Recommendations.}
  \label{fig:dr_agent}
\end{figure*}


\section{Rule-based Search Log Filtering (PySpark)}

In large-scale e-commerce systems, search interaction logs serve as a rich source of user intent signals, behavioral patterns, and optimization opportunities. In this work, we employ a PySpark-based framework to execute a rule-based filtering strategy, enabling systematic identification of high-potential queries for downstream product research.
The filtering pipeline is driven by a set of explicitly defined heuristics derived from business insights and user behavior signals (e.g., engagement patterns, query semantics, and user attributes). These rules are applied at scale using PySpark to efficiently prune the search space and retain only queries that are most likely to benefit from research-driven recommendations.

\subsection{Filtering Criteria}

We apply following filters to the search logs to identify high-potential queries.

\paragraph{1. Queries with No Clicks:}
We filter impressions with zero downstream engagement by applying a predicate on click signals using the \texttt{filter()} transformation. This condition isolates queries where retrieved results failed to generate user interaction, indicating potential intent mismatch.

\paragraph{2. High Affluence Users:}
User affluence is directly available as a feature in the logs. We apply a categorical filter to retain only high-affluence users. This enables prioritization of queries with higher expected conversion impact without requiring external enrichment.

\paragraph{3. Subjective Queries:}
We perform lexical filtering on normalized query text using built-in PySpark string functions.  Queries containing subjective qualifiers (e.g., \textit{best}, \textit{latest}, \textit{top}, \textit{good}, \textit{around}, \textit{under}) are retained via a column-level predicate. These queries typically require reasoning and external knowledge grounding beyond standard retrieval.

\paragraph{4. Vertical-based Filtering (Mobile Category):}
We restrict the dataset to the mobile phones vertical using a product level vertical tag from the logs. This is implemented as a direct \texttt{filter()} condition, avoiding the need for category inference or aggregation. The mobile phones vertical is selected due to its high query volume and rich ecosystem of external knowledge (e.g., specifications, reviews, comparisons). The pipeline remains extensible to other verticals by modifying the filter predicate.

\section{Proposed Agent Architecture}
Fig. \ref{fig:dr_agent} depicts the proposed multi-agent architecture for the product research agent.

\subsection{Design Considerations}


The system is designed using a modular, agent-oriented architecture, where each agent is responsible for a well-defined functional unit. We adopt a centralized orchestrator with specialized agents framework, in which a primary orchestrator agent coordinates the execution flow while delegating domain-specific tasks to specialized sub-agents exposed as callable tools. Each sub-agent writes its output to the shared state, which is subsequently consumed by downstream sub-agents to perform the next stage of processing. A central Supervisor (Orchestrator) agent coordinates the overall execution by invoking sub-agents, managing control flow, and propagating the necessary context across stages. Furthermore, each sub-agent produces structured outputs, enabling deterministic composition of the pipeline and enhancing the interpretability of intermediate results.
A key component of the architecture is a dedicated Review Agent layer integrated at the final stage of the product research pipeline. This layer performs validation and consistency checks over the generated recommendations, thereby improving reliability and adherence to constraints before results are presented to the user.

We also experimented with a Sequential Agents architecture, where agents are invoked in a fixed, predefined order without centralized orchestration. However, empirical evaluation shows that the hierarchical architecture leads to significantly lower instruction-violation error rate. A detailed comparison is presented in the ablation study (Section~\ref{sec: abl}).




\subsection{Orchestrator Agent}
The Orchestrator agent serves as the entry point for all user queries and governs the execution of downstream components. It first leverages the Query Analysis agent to determine whether the query requires research-oriented processing.
Despite upstream rule-based filtering for subjective queries, certain inputs still exhibit specific intent (e.g., “iphone 17 pro new”, “latest vivo y73”), where the target product is already well-defined. For such cases, the orchestrator terminates the pipeline early to avoid unnecessary computation. This distinction is critical, as the system is optimized for exploratory discovery scenarios where multi-step reasoning and external knowledge integration provide significant value.
For queries classified as exploratory, the orchestrator sequentially invokes the Discovery Agent and the Flipkart Search Agent. The product recommendations are then passed to the Review Agent, which performs validation before the final user notification.




\subsection{Query Analysis Agent}

The Query Analysis Agent extracts structured signals from the user query, including intent classification (broad discovery vs. specific), product category (e.g., smartphones, tablets, televisions), and budget constraints. Queries that explicitly specify a product model and variant are classified as \textit{specific}. The extracted signals are returned to the Orchestrator agent. 





\subsection{Discovery Agent}

The Discovery Agent is responsible for generating a high-recall set of candidate products by leveraging external knowledge sources. It is integrated with a web search tool (e.g., Google Search in our implementation) to retrieve heterogeneous content from technology review platforms, editorial blogs, and video-based sources.
The pipeline begins with query expansion, where the input query is reformulated into multiple semantically diverse variants to improve retrieval coverage. This is implemented using prompt-driven transformations that introduce variations in intent (e.g., ranking-focused, comparison-oriented, and feature-specific queries). The expanded query set is then issued to the web search tool, and the retrieved documents are aggregated.
From the retrieved corpus, the agent performs frequency-based and consensus-driven candidate extraction, identifying products that consistently appear across multiple independent sources. This cross-source agreement acts as a weak signal for product relevance and quality.
In addition, the agent extracts temporal signals, particularly product launch timelines, which are critical for queries with recency sensitivity (e.g., “latest”, “new”). For the Indian market, where product availability and launch lag can vary, the agent cross-references multiple sources to infer reliable launch dates.
For each candidate product, the agent generates a structured reasoning narrative that justifies its inclusion. The reasoning incorporates factors such as alignment with query constraints (e.g., price range, brand preferences), feature suitability, and cross-source consensus. To ensure transparency and traceability, the agent also attaches references to the external sources used during retrieval and synthesis.
The final output of the Discovery Agent is a structured candidate set, where each product is associated with metadata including inferred launch date, synthesized reasoning, and supporting source references. This output serves as input to downstream agents in the pipeline.


\subsection{Flipkart Search Agent}

The Flipkart Search Agent grounds the discovered product candidates within the constraints of the Flipkart catalog. It interfaces with an internal \texttt{flipkart\_search} tool to map externally identified products to catalog entries.
The agent invokes \texttt{flipkart\_search} tool where it passes the list of product candidates, user account id and pincode to the \texttt{flipkart\_search} tool to identify corresponding products and retrieve their associated product id.
The product ids are referred to as Flipkart Serial Numbers (FSNs). The tool ensures that each product is available and serviceable 
for the user’s pincode, and also provides personalized price considering user tier, bank offers, and ongoing promotions. To ensure consistency and factual correctness, the agent replaces the specifications generated during the discovery phase with the corresponding attributes retrieved from the Flipkart search tool. This grounding step ensures that all recommendations are aligned with up-to-date catalog data.
Additionally, the agent adapts the reasoning content based on the target communication channel. For instance, messaging platforms like WhatsApp impose character limits requiring concise summaries, whereas email supports more detailed explanations. The agent dynamically adjusts the reasoning format to align with these constraints while preserving clarity and informativeness.


The output of the Flipkart Search Agent is a structured set of product names (as appears on Flipkart), corresponding FSN, product price, product reasoning, and supporting review sources (from Discovery agent).

\subsection{Implementation details of \texttt{flipkart\_search} tool}


The \texttt{flipkart\_search} tool integrates multiple internal APIs, including the Flipkart Search API and Pricing API, to return products that are both serviceable and provide personalized price for a given user.
The tool accepts a list of product names along with user-specific inputs as account ID and pincode. For each product name, it queries the Flipkart Search API to retrieve top-k ranked listings and selects the first listing that is both available and serviceable for the pincode, capturing the corresponding FSN.
The FSN, along with the user account ID, is then passed to the Flipkart Pricing API, which then returns the personalized price with relevant discounts and offers. Note that, the same FSN may yield different prices depending on the user account ID.

Since the search process selects the first serviceable FSN from the top-k results, there is a small possibility of incorrect product matching. Such inconsistencies are subsequently handled by the Review Agent, which filters out incorrect matches.

\subsection{Review Agent}

The Review Agent serves as the final validation layer in the pipeline, ensuring factual correctness, constraint adherence, and overall relevance of the recommended products before they are presented to the user.
This agent performs post-hoc verification through a combination of external re-validation and internal consistency checks. First, it re-validates the product launch dates, using the web search tool to ensure consistency with publicly available sources.
Second, it verifies product-level factual details using the \texttt{get\_fsn\_details} tool. This tool takes FSNs (obtained from the Flipkart Search Agent) as input and returns structured product metadata, including title, description, and detailed specifications. The agent cross-checks all specification claims referenced in the generated reasoning against this metadata to ensure factual alignment and eliminate hallucinated or inconsistent attributes.
In addition to factual verification, the Review Agent enforces query-level constraints. It evaluates each candidate product against the original query requirements, including relevance, budget constraints, category alignment, and feature-specific conditions. 
Products that fail any validation criterion are pruned from the candidate set, while only those satisfying all constraints are marked as verified. The final verified set is then formatted into a structured, templatized response. The Review Agent is implemented as a parallel architecture, wherein independent sub-agents perform specification validation and launch date verification concurrently. This parallelization reduces latency while preserving the robustness of the verification process.

\section{Templated WhatsApp message}

We selected WhatsApp (WA) as the primary notification channel. Due to WhatsApp’s strict message length constraints, the generated reasoning is optimized for brevity while preserving key information. The recommended product name, product price, and agent-generated reasoning are incorporated into a templated WhatsApp message before delivery to the user.
Each recommendation notification includes direct product links, enabling seamless navigation and facilitating faster purchase decisions.


\subsection{Product URL Shortening and UTM Tags for Click Analysis}

Due to the stringent character limitations imposed by WhatsApp message templates, directly embedding full-length product URLs is often impractical, as long URLs increase message length and negatively impact readability and user experience. To address this limitation, we employ an in-house URL shortening service that converts the original product URLs into compact redirect links before embedding them into the outgoing WA messages. The shortened URLs preserve the destination semantics while significantly reducing the number of characters included in the message payload, thereby enabling concise and cleaner message formatting.

In addition to URL shortening, we incorporate campaign-specific UTM (Urchin Tracking Module) parameters into the original product URLs prior to the shortening step. These UTM tags serve as lightweight tracking identifiers that enable downstream analytics and click attribution. Specifically, we introduce day-wise UTM tags that are consistently appended to all product URLs distributed on a given day. This design enables consolidated aggregation of click-through statistics at the campaign-day granularity without requiring product-specific tracking infrastructure. Consequently, all user interactions originating from a particular WA campaign batch can be efficiently grouped and analyzed using standard web analytics pipelines.

The proposed pipeline therefore consists of three sequential stages: (i) generation of the original product URL, (ii) augmentation with campaign-level UTM tracking parameters, and (iii) transformation into a compact shortened URL using the internal URL shortening service. When a user clicks the shortened URL, the redirect service transparently forwards the request to the corresponding original URL containing the embedded UTM parameters. This mechanism enables accurate measurement of key engagement metrics such as click-through rate (CTR), unique clicks, and day-wise campaign performance, while simultaneously maintaining a compact WA message format suitable for large-scale deployment.








\section{Experimental Results}

We executed the product research agent on queries from recent search logs after applying PySpark-based filtering. 
We deployed the WhatsApp (WA) campaign in production over a period of 23 days. For generating the CRM WA messages for each day, we executed an automated end-to-end pipeline. Subsequently, a templated WA message was constructed using the generated recommendations and delivered through the WhatsApp messaging service on the following day.
During the experimental period, a total of 15,061 WA messages were delivered to different users. Each WA message contained two to three product recommendations along with shortened URLs instrumented using day-specific UTM tags for click tracking and attribution analysis. Based on the UTM-tag analysis, we observed a total of 37,258 visits generated from the campaign.

Interestingly, on multiple days, the number of visits significantly exceeded the number of WhatsApp (WA) messages delivered. We attribute this phenomenon to organic message forwarding behavior, where users shared the WA recommendations with other users who subsequently clicked on the embedded product links. This forwarding effect indicates that the recommendations generated by the product research agent were perceived as relevant and useful by users, thereby increasing the effective reach of the campaign beyond the originally targeted audience.

Table \ref{tab:ui} presents the WhatsApp (WA) message read rates and click-through rates (CTR) in comparison with historical WA mobile campaigns within Flipkart. For the AI-agent-driven campaign, the reported metrics are averaged over a 23-day evaluation period.
We observe that the AI-agent-driven campaign achieves a substantially higher CTR compared to the historical campaign baseline. This improvement is driven not only by the increased relevance and personalization of the generated recommendations, but also by the organic forwarding behavior that further amplified user engagement.

In contrast, the relative increase in message read rate is comparatively modest (\textasciitilde 8\%). This discrepancy arises because the read-rate metric can only be computed for users who directly received the WA message. Due to message forwarding, it is not possible to reliably track read rates for secondary recipients who received the messages through forwarded shares. However, because the product URLs contain UTM tags, the CTR metric is able to capture both direct and indirect engagement effects arising from forwarded messages, whereas the read-rate metric reflects only direct message delivery.

\begin{table}[h]
\centering
\begin{tabular}{|l|l|l|}
\hline
\textbf{} & \textbf{\% WA message reads} & \textbf{CTR} \\
\hline
\textbf{AI Agent campaign} &  \textbf{\textasciitilde +8\%} &  \textbf{\textasciitilde +285\%}\\
\hline

\end{tabular}
\caption{Comparison of Historical WhatsApp (WA) Metrics for mobile product campaigns and AI Agent Campaign Interaction Statistics (23-Day Average).}
\label{tab:ui}

\end{table}

The campaign further demonstrated strong business impact in terms of Gross Merchandise Value (GMV). To evaluate downstream conversion behavior, we analyzed order logs over a 15-day period following WhatsApp message delivery. The analysis revealed a substantial overlap between the targeted users, the products recommended through WhatsApp, and the products subsequently purchased in later user sessions. The observed improvements in click-through rates, secondary engagement through message forwarding, and downstream purchase activity collectively highlight the effectiveness of integrating AI agents into enterprise-scale commerce pipelines for user re-engagement.



\subsection{Cost and latency estimates}

We use Google Gemini 2.5 Flash as the underlying LLM. Generating product recommendations for each query involves approximately eight LLM calls across different agents. The average number of input, output, and thinking tokens per query are \textasciitilde 20K, \textasciitilde 4K, and \textasciitilde 3.3K respectively, resulting in an estimated inference cost of \textasciitilde \$0.02-\$0.03 per query. The end-to-end recommendation generation latency is approximately 15--20 seconds. Since the system operates in an offline CRM campaign setting, latency is not a primary concern.

\subsection{Factual correctness metrics}

We evaluate the factual accuracy of generated reasoning through manual annotation by the operations (Ops) team. The evaluation dataset comprises 2,218 product recommendations spanning 730 user queries. Annotators assessed two dimensions: (i) specification accuracy and (ii) launch date accuracy.

For specification accuracy, annotators verified whether the attributes mentioned in the reasoning (e.g., RAM, storage, battery, camera) exactly matched the corresponding product details on Flipkart. A recommendation was labeled as \textit{relevant} only if all listed specifications were correct; any mismatch resulted in an \textit{irrelevant} label. For launch date validation, annotators independently verified the date via google search and compared it against the date stated in the reasoning. Any discrepancy led to the instance being marked as \textit{irrelevant}.

Both specification and launch date are evaluated using accuracy as the primary metric. Table \ref{tab:spac} shows the accuracy metrics for the specs and launch date. The results indicate that incorporating the review agent provides high factual precision by filtering the inconsistencies in generated reasoning.

\begin{table}[h]
\centering
\begin{tabular}{|l|l|l|}
\hline
\textbf{No. of products} & \textbf{spec accuracy} & \textbf{launch date accuracy} \\
\hline
2218 & 99.1 \% & 99.2 \%\\
\hline

\end{tabular}
\caption{Quantitative manual evaluation of factual accuracy}
\label{tab:spac}

\end{table}

\begin{table}[!h]
\centering
\begin{tabular}{|l|l|}
\hline
\textbf{Architecture} & \textbf{Instruction Violation Rate (\%)} \\
\hline
Centralized orchestration & 8.5\%\\
\hline
Sequential Agents &  \textbf{35.4\%} \\
\hline

\end{tabular}
\caption{Effect of the choice of the agent architecture. It is observed that hierarchical supervisor-worker architecture provides superior performance.}
\label{tab:if}

\end{table}

\section{Ablation Study} \label{sec: abl}

We compare a centralized orchestration strategy against a decentralized sequential-agent architecture. In the sequential setup, agents are invoked in a fixed pipeline—query analysis agent $\rightarrow$ discovery agent $\rightarrow$ Flipkart search agent $\rightarrow$ review agent—where each agent consumes the output generated by the previous stage without any centralized coordination. To isolate the effect of architectural design, all prompts, tool interfaces, and configurations are kept identical across both setups.
The evaluation focuses on \textbf{instruction-violation rate} under strict response constraints, motivated by the character limits imposed by messaging platforms such as WhatsApp. As described in Table~\ref{tab:products}, we enforce conditional reasoning augmentation based on query intent. Specifically, queries containing recency cues (e.g., \textit{“latest”}) are expected to include launch-date information in the reasoning, while queries expressing commercial intent (e.g., \textit{“offer”}, \textit{“discount”}) must include discount-related details. In the absence of such cues, no additional reasoning should be included, ensuring concise responses and efficient utilization of the character budget.
We operationalize this evaluation using rule-based checks over the generated outputs. For instance, if a query contains a recency trigger (e.g., \textit{“latest”}) and the response omits launch-date information, it is marked as an error. Conversely, inclusion of such details without the corresponding trigger is also penalized. Similar bidirectional checks are applied for offer- and discount-related cues. The evaluation is conducted over 2.2K product recommendation instances spanning a diverse set of user queries.
Table~\ref{tab:if} reports the instruction-violation rates for both architectures. We observe that the sequential architecture exhibits a substantially higher error rate, particularly in adhering to conditional reasoning requirements such as recency and discount cues. This degradation can be attributed to the progressive transformation of intermediate representations across agents, where critical instruction signals are attenuated or lost due to the absence of a centralized control mechanism.
Our findings are consistent with prior work \cite{kulkarni2026benchmarkingmultiagentllmarchitectures}, which shows that sequential agent pipelines tend to underperform hierarchical architectures for the complex tasks.



\section{Related works}
Recent advances in large language models (LLMs) have led to the emergence of agentic systems, where models are augmented with reasoning, planning, and tool-use capabilities to solve complex tasks \cite{yao2023reactsynergizingreasoningacting}\cite{talebirad2023multiagentcollaborationharnessingpower}\cite{tao2024magisllmbasedmultiagentframework}\cite{trirat2025automlagentmultiagentllmframework}.
Dammu et al. \cite{Dammu2025} have explored LLM-driven agents to address subjective and exploratory queries in e-commerce. They highlights challenges in scenarios like gifting, where user needs are subjective information, and proposes an agentic system leveraging reviews, conversations, and web browsing. 
Huang et al. \cite{huang2024recommenderaiagentintegrating} introduces a hybrid framework where the LLM acts as a reasoning engine while recommender models function as tools. It incorporates mechanisms for handling multi-turn dialogue, improving intent understanding and recommendation quality.
Multi-agent orchestration has been proposed for complex recommendation scenarios. Valentini et al. \cite{valentini2025leveraging} presents a SofAgent for sofa and furniture recommendation. It uses a manager-based architecture coordinating specialized sub-agents for tasks like search and style recommendation. This modular design enables better handling of subjective preferences and product composition in high-involvement domains. Training methodologies for agentic systems have also been explored. Wang et al. \cite{wang2026productresearchtrainingecommercedeep} proposes Multi-Agent Synthetic Trajectory Distillation, where a Supervisor Agent guides a Research Agent through iterative reasoning and tool use. This improves the quality and consistency of generated shopping outputs.
Finally, surveys provide a broader perspective on this space. Peng et al. \cite{Peng2025ASO} categorizes LLM-powered recommender agents into multiple paradigms and analyzes key components such as memory, planning, and interaction. It highlights current challenges and outlines future research directions for agent-based recommendation systems.

\section{Conclusion}
In this work, we presented a scalable end-to-end enterprise framework that bridges search and CRM workflows using AI-driven Product Research Agents for customer re-engagement in large-scale e-commerce environments. The proposed system integrates behavioral analytics, grounded multi-agent reasoning, recommendation validation, and proactive communication channels into a unified production pipeline for exploratory product discovery and customer re-engagement. At the core of the framework is a modular multi-agent architecture coordinated through centralized orchestration, enabling reliable collaboration across specialized agents responsible for intent understanding, external knowledge grounding, candidate retrieval, recommendation generation, and factual validation. Through controlled experimentation, we observed that centralized orchestration substantially improves coordination reliability, factual consistency, and adherence to system-level constraints compared to decentralized sequential agent pipelines.
We further validated the effectiveness of the framework through a live 23-day production-scale WhatsApp campaign for mobile product recommendations. The deployment achieved substantially higher click-through rates (CTR) compared to prior recommendation campaign baselines, while also exhibiting strong evidence of secondary engagement through organic message forwarding and sharing behavior. Beyond engagement improvements, the campaign generated meaningful downstream purchasing activity and GMV impact. Overall, our findings highlight the practical viability of scalable multi-agent AI systems as a unified mechanism for personalized product discovery, CRM-driven re-engagement, and end-to-end customer journey optimization in real-world e-commerce platforms.

\bibliographystyle{ACM-Reference-Format}
\bibliography{sample-base}

@misc{wang2026productresearchtrainingecommercedeep,
      title={ProductResearch: Training E-Commerce Deep Research Agents via Multi-Agent Synthetic Trajectory Distillation}, 
      author={Jiangyuan Wang and Kejun Xiao and Huaipeng Zhao and Tao Luo and Xiaoyi Zeng},
      year={2026},
      eprint={2602.23716},
      archivePrefix={arXiv},
      primaryClass={cs.AI},
      url={https://arxiv.org/abs/2602.23716}, 
}

@Article{Dammu2025,
 author = {Preetam Dammu and Omar Alonso and Barbara Poblete},
 title = {A shopping agent for addressing subjective product needs},
 year = {2025},
 url = {https://www.amazon.science/publications/a-shopping-agent-for-addressing-subjective-product-needs},
}

@misc{kulkarni2026benchmarkingmultiagentllmarchitectures,
      title={Benchmarking Multi-Agent LLM Architectures for Financial Document Processing: A Comparative Study of Orchestration Patterns, Cost-Accuracy Tradeoffs and Production Scaling Strategies}, 
      author={Siddhant Kulkarni and Yukta Kulkarni},
      year={2026},
      eprint={2603.22651},
      archivePrefix={arXiv},
      primaryClass={cs.AI},
      url={https://arxiv.org/abs/2603.22651}, 
}

@misc{huang2024recommenderaiagentintegrating,
      title={Recommender AI Agent: Integrating Large Language Models for Interactive Recommendations}, 
      author={Xu Huang and Jianxun Lian and Yuxuan Lei and Jing Yao and Defu Lian and Xing Xie},
      year={2024},
      eprint={2308.16505},
      archivePrefix={arXiv},
      primaryClass={cs.IR},
      url={https://arxiv.org/abs/2308.16505}, 
}

@article{valentini2025leveraging,
  title={Leveraging LLM-Powered Multi-Agent Systems to Enhance Customer Experience in Complex Product Domains},
  author={Valentini, Marco and Ferrara, Antonio and Di Noia, Tommaso and Illuzzi, Giuseppe and Colacicco, Pierangelo},
  year={2025}
}

@article{Peng2025ASO,
  title={A Survey on LLM-powered Agents for Recommender Systems},
  author={Qiyao Peng and Hongtao Liu and Hua Huang and Qing Yang and Minglai Shao},
  journal={ArXiv},
  year={2025},
  volume={abs/2502.10050},
  url={https://api.semanticscholar.org/CorpusID:276395083}
}

@misc{yao2023reactsynergizingreasoningacting,
      title={ReAct: Synergizing Reasoning and Acting in Language Models}, 
      author={Shunyu Yao and Jeffrey Zhao and Dian Yu and Nan Du and Izhak Shafran and Karthik Narasimhan and Yuan Cao},
      year={2023},
      eprint={2210.03629},
      archivePrefix={arXiv},
      primaryClass={cs.CL},
      url={https://arxiv.org/abs/2210.03629}, 
}

@misc{talebirad2023multiagentcollaborationharnessingpower,
      title={Multi-Agent Collaboration: Harnessing the Power of Intelligent LLM Agents}, 
      author={Yashar Talebirad and Amirhossein Nadiri},
      year={2023},
      eprint={2306.03314},
      archivePrefix={arXiv},
      primaryClass={cs.AI},
      url={https://arxiv.org/abs/2306.03314}, 
}

@misc{tao2024magisllmbasedmultiagentframework,
      title={MAGIS: LLM-Based Multi-Agent Framework for GitHub Issue Resolution}, 
      author={Wei Tao and Yucheng Zhou and Yanlin Wang and Wenqiang Zhang and Hongyu Zhang and Yu Cheng},
      year={2024},
      eprint={2403.17927},
      archivePrefix={arXiv},
      primaryClass={cs.SE},
      url={https://arxiv.org/abs/2403.17927}, 
}

@misc{trirat2025automlagentmultiagentllmframework,
      title={AutoML-Agent: A Multi-Agent LLM Framework for Full-Pipeline AutoML}, 
      author={Patara Trirat and Wonyong Jeong and Sung Ju Hwang},
      year={2025},
      eprint={2410.02958},
      archivePrefix={arXiv},
      primaryClass={cs.LG},
      url={https://arxiv.org/abs/2410.02958}, 
}

\appendix





\end{document}